\documentclass[sigconf,natbib=false]{acmart}

\AtBeginDocument{%
  }

\usepackage{amsmath}
\usepackage{float}
\usepackage{caption}
\usepackage{dblfloatfix}
\usepackage{xcolor}

\newtheorem{remark}{{Remark}}

\newcommand{\marginXW}[1]{\marginpar{\color{purple}\tiny\ttfamily Xuan: #1}}

\setcopyright{acmlicensed}
\copyrightyear{2024}
\acmYear{2024}
\acmDOI{}

\acmConference[ACM-BCB]{ACM Conference on Bioinformatics, Computational Biology, and Health Informatics}{Nov. 22--25,
  2024}{Shenzhen, China}
\acmBooktitle{ACM Conference on Bioinformatics, Computational Biology, and Health Informatics (ACM-BCB)}

\RequirePackage[
  datamodel=acmdatamodel,
  style=acmnumeric,
  ]{biblatex}

\begin{document}

\title{Causality-based Subject and Task Fingerprints using fMRI Time-series
Data}








\author{Dachuan Song}
\affiliation{%
  \institution{Department of Electrical and Computer Engineering, George Mason University}
  \city{Fairfax}
  \state{Virginia}
  \country{USA}
}

\author{Li Shen}
\affiliation{%
  \institution{Department of Biostatistics, Epidemiology and Informatics, Perelman School of Medicine, University of Pennsylvania}
  \city{Philadelphia}
  \state{Pennsylvania}
  \country{USA}
}

\author{Duy Duong-Tran}
\affiliation{%
  \institution{Department of Biostatistics, Epidemiology and Informatics, Perelman School of Medicine, University of Pennsylvania}
  \city{Philadelphia}
  \state{Pennsylvania}
  \country{USA}
}
\affiliation{%
  \institution{Department of Mathematics, United States Naval Academy}
  \city{Annapolis}
  \state{Maryland}
  \country{USA}
}

\author{Xuan Wang}
\affiliation{%
  \institution{Department of Electrical and Computer Engineering, George Mason University}
  \city{Fairfax}
  \state{Virginia}
  \country{USA}
}

\renewcommand{\shortauthors}{}

\begin{abstract}
Recently, there has been a revived interest in system neuroscience causation models due to their unique capability to unravel complex relationships in multi-scale brain networks. In this paper, our goal is to verify the feasibility and effectiveness of using a causality-based approach for fMRI fingerprinting. Specifically, we propose an innovative method that utilizes the causal dynamics activities of the brain to identify the unique cognitive patterns of individuals (e.g., subject fingerprint) and fMRI tasks (e.g., task fingerprint). The key novelty of our approach stems from the development of a two-timescale linear state-space model to extract `spatio-temporal' (aka causal) signatures from an individual's fMRI time series data. To the best of our knowledge, we pioneer and subsequently quantify, in this paper, the concept of `causal fingerprint.' Our method is well-separated from other fingerprint studies as we quantify fingerprints from a cause-and-effect perspective, which are then incorporated with a modal decomposition and projection method to perform subject identification and a GNN-based (Graph Neural Network) model to perform task identification.
Finally, we show that the experimental results and comparisons with non-causality-based methods demonstrate the effectiveness of the proposed methods. We visualize the obtained causal signatures and discuss their biological relevance in light of the existing understanding of brain functionalities.
Collectively, our work paves the way for further studies on causal fingerprints with potential applications in both healthy controls and neurodegenerative diseases. 
\end{abstract}





\maketitle
\section{Introduction}
The advancements in brain imaging technologies, such as Functional Magnetic Resonance Imaging (fMRI) of Blood-Oxygen-Level-Dependent (BOLD) signals, have provided quantified measures to capture brain activities, leading to a significant boost in neuroscience research~\cite{logothetis2002neural}. Among the various ways of brain modeling, there has been a revamped and growing interest in causation models~\cite{kiebel2007dynamic,ross2024causation}, due to their unique focus on unraveling the complex interactive relationships in multi-scale brain networks. These models offer pathways to understand the internal operating mechanisms of the brain~\cite{mannino2015foundational}; elucidate how such mechanisms lead to human cognitive functions and behavioral responses~\cite{rigoux2015dynamic,duong2021morphospace,garai2023mining}; and open up possibilities for `manipulating' brain activities~\cite{gu2015controllability,amico2021toward} to enhance regular functionalities and mitigate disorders \cite{xu2022consistency}.
Enlightened by these, this paper explores and verifies the effectiveness of using causation models in the context of fMRI fingerprinting. The goal is to utilize causal signatures derived from fMRI time-series data to determine the tester's identity (subject fingerprint) and the fMRI tasks they are performing (task fingerprint).
To achieve this, a primary challenge is the diverse subject/task variations, compounded by the limited data available for each subject-task pair~\cite{SM-SM-VI-DI-BE}.

The concept of the fMRI fingerprint \cite{abbas2020geff,abbas2023tangent,amico2021toward,duong2021morphospace,chiem2022improving,duong2024homological,duong2024principled,duong2024preserving,duong2024principled_m} can be framed as a classification problem. Given the constraints of data availability, the explainability and interoperability of any developed algorithm are critical for ensuring its robustness. For instance, in the field of person re-identification (re-ID) using visual inputs~\cite{YM-SJ-LG-XT-SL:21}, leveraging augmented features like gender, color, and textures can significantly enhance accuracy. In contrast, raw brain signals present inherent challenges in their explanation or translation into meaningful features.
To tackle these challenges, the functional connectome (FC) has emerged as a promising method~\cite{SM-SM-VI-DI-BE}. FC measures the correlations between different brain regions during rest or task conditions and has been successful in identifying subject-specific signatures and classifying various cognitive states and disorders~\cite{CB-ZG-ZA-XL-HW-SJ-WT-CV-WY:21,FES-SX-SD-RMD-HJ:15}. Other brain modeling methods in the literature include Independent Component Analysis (ICA), which decomposes fMRI data into spatially independent components representing distinct brain networks~\cite{CA-VD-AD-TU-PE,MC-MJ-MA-SC-BR}. ICA is particularly useful for uncovering resting-state networks (RSNs) and understanding their roles in cognitive functions and pathologies~\cite{BE-CF-DE-MA-DE,SM-SM-FO-PT-MI:2009,VD-MP-PO-HE:2010}.
Despite the successes of FC and ICA, they also have limitations: they do not fully explore the brain's dynamic nature, lack temporal resolution for the interactions among different regions, and do not capture the directionality of these relationships~\cite{CB-ZG-ZA-XL-HW-SJ-WT-CV-WY:21,UM-SL-DD-WX:23,abbas2023tangent,duong2024homological}. All of these are essential for understanding the \textit{cause-and-effect} interactions within the brain's network.

In contrast to non-causal methods, causality-based approaches such as Granger causality and Dynamic Causal Modeling (DCM) offer alternative insights into brain interaction patterns, which however, are less explored in the context of fingerprinting. Causality in fMRI involves analyzing directional influences among brain regions over time, providing insights into the brain's dynamic and interactive modes. Granger causality, for instance, determines whether one time series can predict another, revealing potential causal relationships~\cite{GR-CW,B-AB-B-L,R-A-F-E-G}. DCM, using a Bayesian framework, models interactions among neural states using bi-linear functions to provide insights into how different brain regions influence each other over time~\cite{FR-KJ-HA-L-P-W,S-KE-P-WD-D}. In recent years, modifications of these models have been developed, such as introducing threshold structures~\cite{wang2021data} to represent activation saturation, and multifactorial dynamics \cite{iturria2017multifactorial} to represent multi-time-scale time-varying systems. 

Given these considerations, our study poses an intuitive but unanswered question: can causality be utilized as a definitive signature of brain activity for performing subject and task fingerprinting? By leveraging the temporal and directional information inherent in causality models, this paper aims to develop robust and accurate fMRI fingerprinting methods for both subject and task identification.

\textit{Statement of contribution:} In this paper, our result provides a solid `YES' answer to the feasibility and effectiveness of utilizing brain causal signatures for subject and task fingerprinting. As such, we pose the quest of human identifiability, via brain fMRI imaging modality, through the brain hidden causal signatures. Here, we would like to emphasize that finding and quantifying causal signatures in the context of fingerprint is a more challenging, specific sub-problem, compared to generalized (e.g., non-causal) fingerprint quests. As such, conventional fingerprint investigations \cite{abbas2020geff,abbas2023tangent,amico2021toward,duong2021morphospace,chiem2022improving,duong2024homological,duong2024principled, FES-SX-SD-RMD-HJ:15, amico2018quest} are correlation-based, hence, incapable of capturing cause-and-effect cognitive signatures among individuals. 
The key novelty of our approach stems from the data-driven reconstruction of a two-timescale linear state-space model to extract `spatio-temporal' (aka causal) signatures from fMRI time series data. 
We observe that these signatures exhibit characteristics of both control and network systems. This has inspired us to associate it with the concept of dynamic mode~\cite{proctor2016dynamic} in control theory, leading to the development of a modal decomposition and projection method for robust subject fingerprint, and a GNN-based (Graph Neural Network) model that leverages the graph representation of the network system to perform accurate task fingerprint.
Although these methodologies have been applied in other contexts, to our knowledge, this work is the first to integrate them into the `causal fingerprint' concept, which is a quantifiable measure describing the degree to which a subject or an fMRI task can be identified from their unique causal cognitive patterns. 
The visualization of these causal signatures allows us to discuss the corresponding biological mechanism and connect them with the existing understanding of brain functionalities.
Compared with other fingerprint explorations that do not have a causation interpretation, the inherent explainability of causality not only allows us to pursue more sophisticated fingerprint quests but also paves the way for further studies on causal fingerprints in both healthy controls and neurodegenerative diseases.

\section{Methods: Causal Fingerprint}



We start this section by formally defining the concept of causal fingerprint {(later referred to as fingerprint, for simplicity.)}
\begin{definition}
    Causal fingerprint is the degree to which a subject or an fMRI task can be identified from a labeled database particularly based on the subjects' or fMRI tasks' unique \textit{cause-and-effect} cognitive signatures.
\end{definition}
To verify the feasibility and effectiveness of using causal signatures for fingerprinting, our method first introduces a two-time scale state-space model to capture these causal signatures. These signatures are then combined with a modal decomposition and projection method for subject fingerprinting and with a GNN model for task fingerprinting.

\noindent \textbf{Notations.} Throughout this section, suppose we are given an fMRI data set with $\mathcal{S}$ representing the set of all subjects, and $\mathcal{K}$ representing the set of all tasks performed by the subjects. 
Let $d_{s,k}\in\mathbb{R}^{p\times T}$ represent the time-series recording of a subject $s\in\mathcal{S}$ performing a fMRI task $k\in\mathcal{K}$. 
The dimension $p$ is the number of brain parcellations and $T$ is the testing duration. Let $d_{s,k}(t)\in\mathbb{R}^{p\times T}$ represent the $t^{th}$ column of $d_{s,k}$. Let $\text{vec}(\cdot)$ represent the vectorization (stacking all columns) of a matrix into a single-column vector. 



\subsection{A Two-timescale State-space Model for Causal Dynamics}\label{sec_model}
\begin{figure*}[t]
\centering
\includegraphics[width=.7\textwidth]{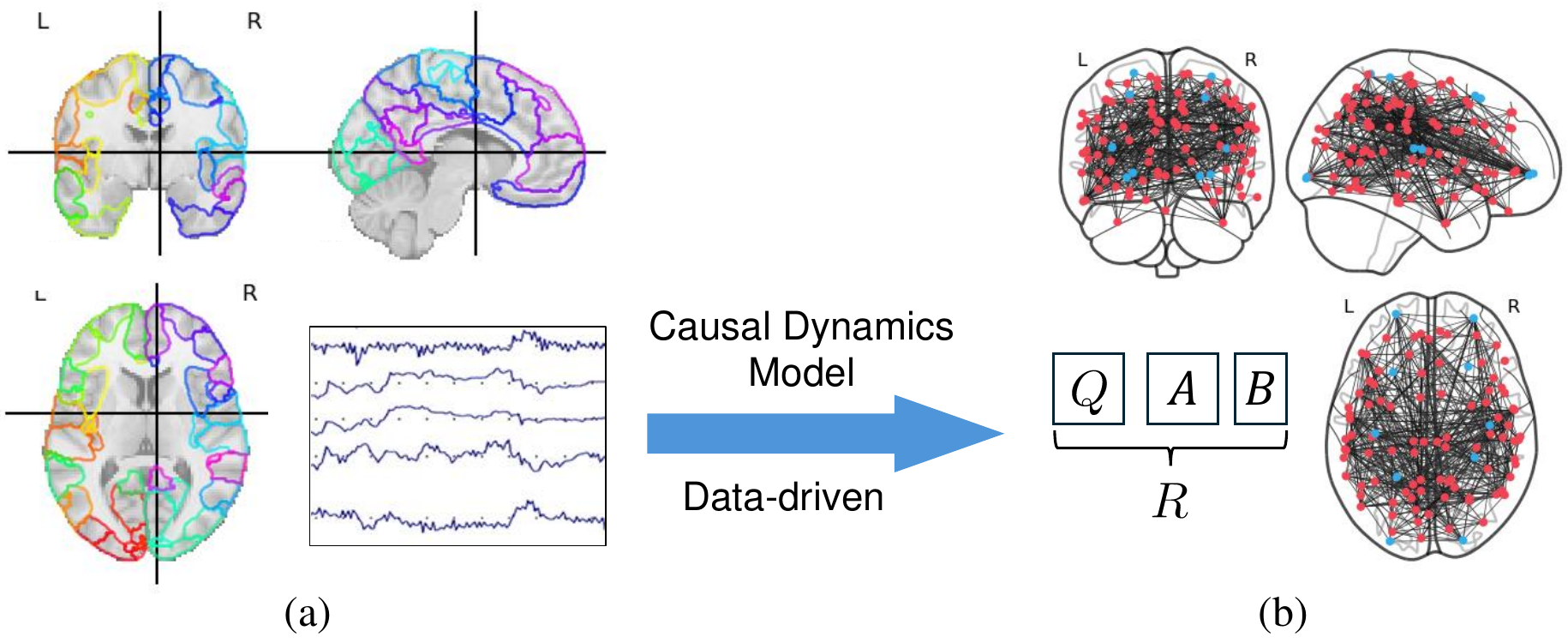} 
\caption{\textbf{(a)} Brain Activity (time-series data) captured using Schaefer Parcellation. \textbf{(b)} Causal Dynamics modeling and data-driven parameter identification.}
\label{fig:brain_connect}
\end{figure*}
To capture the variability of subjects performing different fMRI tasks, we present a new approach based on a \textit{two-timescale} linear state-space model. Its novelty lies in the capability to disentangle \textit{spatio-temporal} signatures of large-scale brain networks, through the data-driven reconstruction of brain causal dynamics from the fMRI time-series data $d_{s,k}$. 
Since this modeling approach applies to any subject for any task, in this subsection, we omit the subscripts ${s,k}$ for simplicity.
\\Consider a discrete-time state-space model with two timescales:
\begin{align}\label{eq_model}
    x(t) = Qx(t) + Ax(t-1) + Bu(t-1)
\end{align}
Here, we divide $p$ brain parcellations into two subsets: system states and inputs, denoted by $x(t)\in\mathbb{R}^m $ and $ u(t)\in\mathbb{R}^n $, $m+n=p$, with each entry of the vectors encapsulate the activity of a specific brain region at time $t$. The matrices $ A\in\mathbb{R}^{m\times m} $, $ B\in\mathbb{R}^{m\times n} $, and $ Q\in\mathbb{R}^{m\times m} $ encode the cause-and-effect relations among different brain regions that evolve over time. From the temporal aspect, we model the system with two timescales, where matrices $A$ and  $B$ capture the slower dynamical evolution of the system, describing how its current states are impacted by its states/inputs from the previous timestep; $Q$ captures the fast interaction that happens concurrently among brain regions. From a spatial aspect, $B$ represents the excitation/inhibition relation between inputs and system states; the off-diagonal entries of $A$ and $Q$ characterize the interdependent excitation/inhibition relation among system states; the diagonal entries of $A$ characterize the intrinsic decay rate of each state from the previous step to the current time-step; the diagonal entries of $Q$ are forced as zeros to avoid a trivial self-mapping for each state.

Based on model \eqref{eq_model}, we employ a data-driven approach to identify its parameters $A$, $B$, and $Q$ from fMRI time-series data. Specifically, assuming the brain parcellations are categorized into systems states and inputs, and the fMRI data sample for a particular subject performing a specific task is recorded as $d(t)=\{x(t), u(t)\}$, for $t\in\{0,\cdots, T\}$. Since the sample pairs should satisfy equation \eqref{eq_model}, if defining $X^{1:T}=\begin{bmatrix}
    x(1)& x(2)&\cdots x(T)
\end{bmatrix}$ and $U^{0:T-1}=\begin{bmatrix}
    u(0)& u(1)&\cdots u(T-1)
\end{bmatrix}$, 
a compact representation of the dynamics follows
$$X^{1:T}=QX^{1:T} + AX^{0:T-1} + BU^{0:T-1}$$
To determine the model parameters $A$, $B$, and $Q$ in the presence of model imperfections and measurement noise, we can formulate the generalized least squares problem:
\begin{align}\label{eq_matrix_id}
       \underset{A,B,Q}{\arg\min}\quad    &\left\|(Q-I)X^{1:T} + AX^{0:T-1} + BU^{0:T-1}\right\|_F,\\\nonumber
     & \text{subject to}\quad Q_{ii}=0,~\forall i\in\mathbf{m} 
\end{align}
where $\|\cdot\|_F$ denotes the Frobenius norm \cite{peng2016connections} to minimize the residue in \eqref{eq_matrix_id}.
By solving \eqref{eq_matrix_id}, we transform a complex fMRI time-series data $d(t)=\{x(t), u(t)\}$ into a structured and meaningful representation ${R}= [Q~A~B]\in\mathbb{R}^{m\times(2m+n)}$ of brain dynamics signature, as visualized in Fig. \ref{fig:brain_connect} .
This transformation not only makes the data more accessible for subsequent analysis but also reveals the patterns of `spatio-temporal’ causality relation that underpin cognitive processes.
In the following, we will build on $R$ to perform subject and task fingerprints.

\begin{remark}\label{RM_1}
Compared with the state-of-the-art FC (functional connectome) representation for fingerprints, the main difference of the proposed causal signature lies in its capability to describe the directed (which area inhibits or excites another area) and temporal (how earlier activities impact subsequent activities) interaction patterns among different brain regions. In contrast, the FC method only describes undirected and concurrent relationships between activities of every pair-wise brain region.

Regarding the model choice, we also remark that this paper uses a linear state-space model \eqref{eq_model}, which is a relatively simple causal representation compared with the diverse literature that introduces more complex causal models with non-linear structures. For example, the bi-linear dynamics in \cite{FR-KJ-HA-L-P-W}, threshold dynamics in \cite{wang2021data}, and multifactorial time-varying dynamics in \cite{iturria2017multifactorial}. Nevertheless, we argue that the use of a simpler model does not undermine the results of the paper; instead, it provides strong evidence for the effectiveness of cause-and-effect signatures for fingerprinting purposes. Simpler models generally have mild requirements on data richness. It will be demonstrated in Experiments that our approach can use low-resolution fMRI data (Schaefer-100) to achieve comparable accuracy to other non-causal-based (FC, end-to-end learning) approaches.
\end{remark}

\subsection{Subject Causal Fingerprint via Modal Decomposition and Projection}\label{sec_subjetid}
Given fMRI time-series data $d_{s,k}$ for any subject-task pair, the model in Sec. \ref{sec_model} allows us to obtain a representation $R_{s,k}$ that extracts its temporal and spatial signatures. In the following, we incorporate these signatures with a state-space modal decomposition~\cite{dahleh2004lectures} and projection method to perform subject fingerprinting. 


For any fMRI task $k\in\mathcal{K}$, we build a labeled database:
$$\widehat{\mathcal{R}}_k=\{R_{s,k}^D\}, \quad s\in\mathcal{S}$$ 
where the superscript $D$ represents labeled data with known subject indices $s$.  We make a necessary assumption about the data set: each subject-task pair has at least two recordings. This allows us to use one recording to construct the database $\widehat{\mathcal{R}}_k$ and reserve the others as query data for testing purposes.
Given a query whose causal signature is $R_{k}^Q$ with an unknown identity, our {goal} is to find the label $s$ such that $R^{D}_{s,k}$ is most \textit{similar} to $R^Q_{k}$.

The key challenge for subject causal fingerprinting lies in its one-shot nature, meaning that only one sample per subject is available to determine their identity from a large number of candidates. This challenge is similar to those faced in visual image-based subject identification, which can be addressed by introducing explainable augmented features such as gender, color, and textures to improve algorithmic accuracy and robustness~\cite{YM-SJ-LG-XT-SL:21}. However, a similar approach is not directly transferable to brain images because raw fMRI time-series data is highly complex and lacks explainable features.
In contrast, the causal signature $R= [Q~A~B]$ are matrices of a dynamic system, which has a unique control theoretic interpretation, allowing us to use its \textit{dynamic modes}~\cite[Chapter 12]{dahleh2004lectures} as augmented features to develop a new method for robust subject fingerprinting.

Specifically, dynamic modes are obtained by the following \textit{modal decomposition and projection}. Let $Q=\bar{T}^{-1}\bar{\Lambda} ~\bar{T}$, where columns of $\bar{T}$ are right eigenvectors of $Q$; $\bar{\Lambda}$ is the diagonalized matrices corresponding to eigenvectors (Jordan matrices if not diagonalizable~\cite{horn2012matrix}). Then by left multiplying $\bar{T}^{-1}$ to model \eqref{eq_model}, it can be rewritten as 
\begin{align}\label{eq_model2}
    \bar{x}(t) = \bar\Lambda \bar{x}(t) + \bar{A}\bar{x}(t-1) + \bar{B}u(t-1)
\end{align}
where $\bar{x}(t)=\bar{T}^{-1}{x}(t)$,  $\bar{A}=\bar{T}^{-1}A~\bar{T}$ and $\bar{B}=\bar{T}^{-1}B$. Since $\bar{\Lambda}$ is diagonal, it describes how each entry of system state $\bar{x}(t)$ evolves in a fully decoupled \textit{component-wise} manner. These components, known as invariant dynamic modes, are embedded in matrix $\bar{T}$ as it projects  ${x}(t)$ to the new coordinates $\bar{x}(t)$. Similarly, let $A=\widehat{T}^{-1}\widehat{\Lambda} ~\widehat{T}$. By left multiplying $\widehat{T}^{-1}$ to \eqref{eq_model}, one has 
\begin{align}\label{eq_model3}
    \widehat{x}(t) = \widehat\Lambda \widehat{x}(t-1) + \widehat{Q}\widehat{x}(t) + \widehat{B}u(t-1)
\end{align}
where $\widehat{x}(t)=\widehat{T}^{-1}{x}(t)$,  $\widehat{Q}=\widehat{T}^{-1}Q~\widehat{T}$ and $\widehat{B}=\widehat{T}^{-1}B$. 
Equation \eqref{eq_model3} describes the dynamics of $\widehat{x}(t)$ in a \textit{component-wise} manner.
The difference between \eqref{eq_model2} and \eqref{eq_model3} arises from our two-time scale model: $\bar\Lambda$ and $\bar{T}$ are computed from $Q$, representing modes associated with the concurrent interaction among brain regions; while $\widehat\Lambda$ and $\widehat{T}$ are computed from $A$, representing modes associated with the slower dynamical evolution of the system.

Since the columns of projection matrices $\bar{T}$ and $\widehat{T}$ embed dynamic modes of \eqref{eq_model}, we use them as augmented features for subjects to perform fingerprinting. We determine the identity of query data by its alignment with database subject labels in terms of their dynamic modes: 
\begin{align}\label{eq_dist}
\underset{s}{\arg\min}\quad\text{Dist}(\{\bar{T},\widehat{T}\}_{R_{k}^Q},~\{\bar{T},\widehat{T}\}_{R_{s,k}^D})
\end{align}
where the distance is measured by the permutations of pairwise vector similarity (Euclidean dot product) of the columns in $\bar{T}$ and $\widehat{T}$. Note that the matrices $\bar\Lambda$ and $\widehat\Lambda$ quantify the significance of each mode in the dynamics. In our approach, we consider all dynamic modes of the system to be equally important, so \eqref{eq_dist} only involves $\{\bar{T},\widehat{T}\}$ and does not consider $\bar\Lambda$ and $\widehat\Lambda$. Furthermore, this distinguishes our approach from similar algebraic projection methods such as PCA (Principal component analysis), which aims to reduce the system's dimension by preserving only the most significant components. In contrast, \eqref{eq_dist} considers the full-dimensional space spanned by all system's dynamic modes. Such treatment builds on our assumption that when different subjects perform the same fMRI task, their major dynamic modes might be similar, but vary in certain minor dynamic modes.

\subsection{Task Causal Fingerprint via Graph Neural Network}
\begin{figure}[b]
\centering
\includegraphics[width=.48\textwidth]{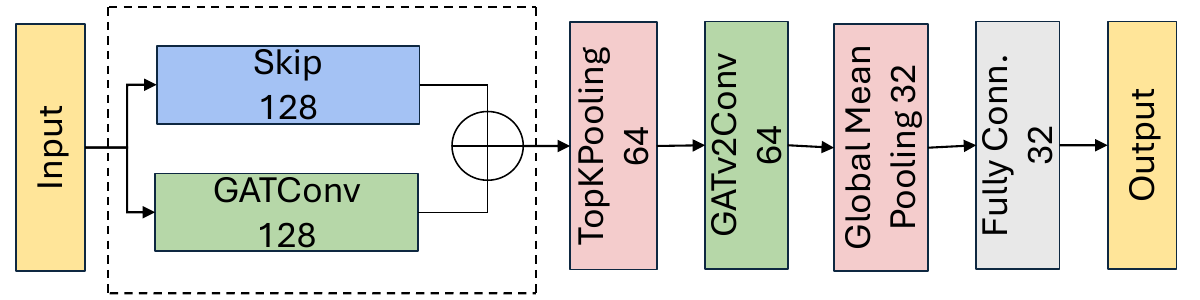} 
\caption{Five-Layer GNN for Task Fingerprinting.}
\label{fig:model_architecture}
\end{figure}

Unlike subject identification, where the key challenge lies in the large number of subjects and a limited dataset for each subject, task identification involves recognizing a task from a limited number of candidates, each with a sufficiently abundant dataset (equal to the total number of subjects). However, although task identification benefits from richer data, it faces inherent challenges due to variations in subjects' brain structures and brain function realizations. Consequently, methods similar to Sec. \ref{sec_subjetid} based on pre-defined algebraic operators are no longer sufficient. 

In this subsection, we instead achieve causal-based task fingerprinting by incorporating the extracted signatures $R_{s,k}$ in Sec. \ref{sec_model}, which encode a graph-theoretic representation of brain networks, into a GNN-based learning model.
To explain our method, we construct a labeled database: $$\widehat{\mathcal{R}}=\{R^{LR}_{s,k}\}, \quad s\in\mathcal{S}, \text{ and } k\in\mathcal{K}$$ 
where $\mathcal{S}$ and $\mathcal{K}$ are subject sets and fMRI task sets, respectively. 
Given a query $R^Q$ data whose identity and task both being unknown, our goal is to identify the task index $(k)$ of the query from the database $\widehat{\mathcal{R}}$.
A fundamental characteristic of GNN is to represent input data into a graph-like structure~\cite{khemani2024review}. This has an intrinsic correspondence with our data matrices $R= [Q~A~B]$, where from a spatial aspect, each row of the matrix describes how a particular brain region is impacted by other brain regions and external inputs. Thus, we set our GNN model $\mathcal{G}(\mathcal{V},\mathcal{E})$ with $\mathcal{V}$ representing the nodes and $|\mathcal{V}|=m$. The feature of each node is a vector of $\mathbb{R}^{(2m+n)}$, associated with each row of ${R}$. 
The normalized $A$ matrix in $R$ is used as an adjacency matrix to initialize edge features $\mathcal{E}$.


Based on the described input structure, our GNN model has a five-layer structure designed to optimize performance in processing fMRI data for task classification. The model begins with a Graph Attention Network (GATConv) layer that dynamically weights the importance of adjacent nodes based on their features. It enhances node feature representation by focusing on the most relevant connections~\cite{velickovic2017graph}. In parallel, skip connections are included to retain original node feature information, preventing information loss and enabling robust learning~\cite{xu2021optimization}. 
The third layer, a TopKPooling layer, selectively prunes the graph to retain only the most significant nodes, reducing computational complexity and focusing on the most informative parts of the graph. This enhances model efficiency and effectiveness by highlighting critical features for subsequent layers. After pooling, a third GAT layer (GATv2Conv) projects the pruned node features into a more discriminative space, which refines node features to make them more distinguishable for the final classification task. 
The fourth layer is a global mean pooling layer which aggregates refined node features across the entire graph, producing a graph-level representation that summarizes all node information into a single vector. This vector captures the overall graph structure and features, making it suitable for the final classification step. Finally, a fully connected linear layer transforms the aggregated graph-level representation into the final classification, mapping the high-dimensional feature vector into specific classes representing different fMRI tasks. This layer enables the model to make accurate predictions based on the input fMRI data. 
The model structure is illustrated in Figure \ref{fig:model_architecture}.



The model is trained with data batches in $\widehat{\mathcal{R}}$ in a supervised learning fashion. The rows of $R^D_{s,t}$ are used as model inputs to each node, and $t$ is used as the fMRI-task label. The training loss is a mismatch of the model output compared with the true label, and the optimizer incorporates weight decay (L2 regularization) to prevent overfitting. This choice is motivated by its adaptive learning rate adjustment capabilities and the inclusion of weight decay to penalize large weights, which helps in improving the generalization of the model. 

\section{Results}

This section uses real-world datasets to verify the effectiveness of the proposed causal-based fingerprints for both subject and task fingerprints.

\subsection{{Dataset and Modeling.}} 

\begin{figure}[h]
\centering
\includegraphics[width=.5\textwidth]
{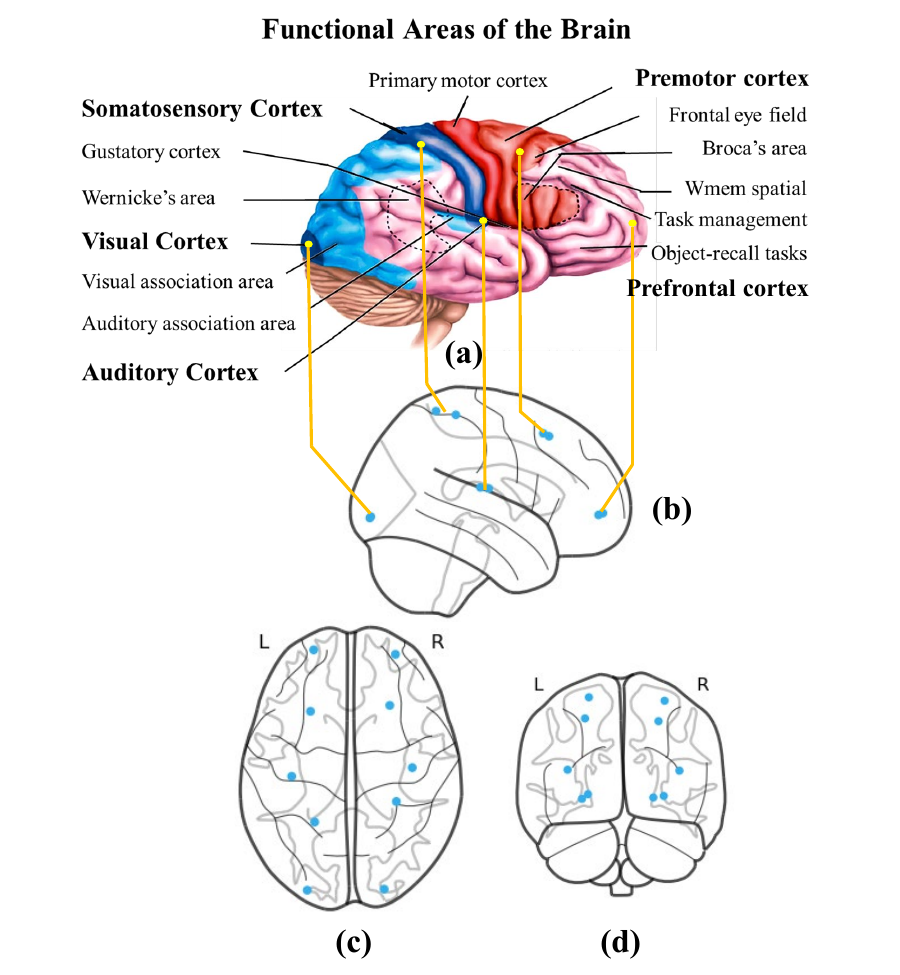}
\caption{(\textbf{a}) Functional Areas of the Brain. (\textbf{b-d}) Ten Schaefer Parcellation Areas (blue dots) that are Chosen as System Inputs: Two areas are associated with the prefrontal cortex; Two areas are associated with the pre-motor cortex; Two areas are associated with the somatic cortex; Two areas are associated with the vision cortex; Two areas are associated with the hearing cortex. We assume these are areas likely to `initiate' certain brain activities. \vspace{-1em}}
\label{fig:brain_map}
\end{figure}

We use the dataset from the Human Connectome Project (HCP)~\cite{van2013wu}, which consists of fMRI time-series signals collected from 391 unrelated subjects\footnote{Unrelated subjects eliminate potential genetic confounders.}. Each subject participated in two resting-state sessions and seven distinct fMRI tasks: emotional response (Emot), gambling (Gamb), language processing (Lang), motor function (Moto), relational processing (Rela), social cognition (Soci), and working memory (WMem). Each subject-task pair has two recordings, scanned in LR (left to right) or RL (right to left) patterns. 
The dataset employs Schaefer parcellation-100~\cite{schaefer2018local}, which divides the brain into $p=100$ distinct regions. Time-series data for each parcellation were collected every 720 milliseconds. 
Echoing Remark 1, although there exists higher resolution data, the Schaefer-100 resolution used here is sufficient for us to explore the extent to which the causal signatures of the brain can be used to determine a subject's identity or fMRI task, and our results have proven this idea. 
Corresponding to model \eqref{eq_model}, among the 100 parcellations, we choose $n=10$ as input nodes, their locations are associated with the premotor and sensory areas of the brain~\cite{hardwick2015multimodal} (as detailed in Figure \ref{fig:brain_map}). Specifically, we choose two areas associated with the prefrontal cortex, two areas associated with the premotor cortex, two areas associated with the somatosensory cortex, two areas associated with the visual cortex, and two areas associated with the auditory cortex. These areas are likely critical for initiating certain brain activities. The reason for selecting these areas is their crucial role in performing cognitive and sensory tasks, including task management, object recall, spatial tasks, sensory processing, visual processing, and auditory processing. This selection is based on the significance of these brain regions in multimodal integration functions, as detailed in the literature \cite{hardwick2015multimodal}. The remaining $m=90$ regions are chosen as system states.

\begin{figure*}[t]
\centering
\includegraphics[width=\textwidth]{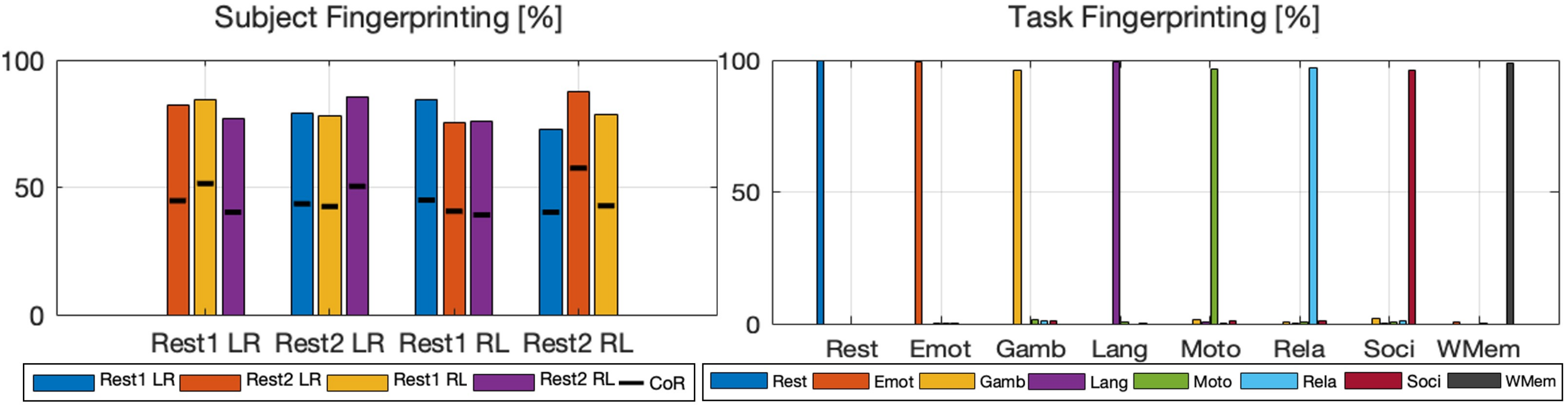}
\caption{Left: Subject Fingerprinting based on CM+MD\&P: The identification accuracy is compared with (black line) using the FC+CoR method.
Right: Task Fingerprinting based on CM+GNN: The figure shows the correct/incorrect classification for the same/different fMRI tasks. 
}
\label{fig:subject_task-accuracy}
\end{figure*}

\begin{table*}[b]
\centering
\caption{Subject Fingerprinting Using Multiple Approaches}
\label{tab:my_label}
{%
\begin{tabular}{|l|c|c|c|c|c|c|c|c|}
\hline
\textbf{Method} & \textbf{CM+MD\&P} & {CM+CoR} & {CM+FN} & {CM+GNN} & {FC+CoR} & {FC+MD\&P} & {FC+FN} & {FC+GNN} \\
\hline
Rest1 LR & \textbf{81.245}\% & 14.834\% &42.251\%& $<5\%$ & 45.439\% & 69.736\% &33.952\%& $<5\%$ \\
Rest2 LR & \textbf{80.904}\% & 14.919\% &47.175\%& $<5\%$ & 45.436\% & 70.929\% &37.574\%& $<5\%$ \\
Rest1 RL & \textbf{78.517}\% & 15.689\% &41.333\%& $<5\%$ & 41.603\% & 67.945\% &38.698\%& $<5\%$ \\
Rest2 RL & \textbf{79.710}\% & 14.834\% &43.463\%& $<5\%$ & 46.803\% & 67.775\% &36.737\%& $<5\%$ \\
\hline
\end{tabular}
}
\begin{tabular}{@{}lll@{}}
CM: Causal Dynamics Model & MD\&P: Modal Decomposition and Projection & CoR: Correlation \\
FC: Functional Connectivity & FN: Frobenius norm & GNN: Graph Neural Network\\
\end{tabular}
\end{table*}
\begin{table*}[b]
\centering
\caption{Task Fingerprinting Accuracy Using Multiple Approaches}
\label{tab:my_label2}
{%
\begin{tabular}{|l|c|c|c|c|c|c|c|c|}
\hline
\textbf{Method} & \textbf{CM+GNN} & {CM+RF} & {CM+SVM} & {FC+GNN} & {Raw+GNN}& {Raw+DNN} & {Raw+DBRNN} & {Raw+BAnD} \\
\hline
Rest & \textbf{100}\% & \textbf{100}\% & 92.442\%  & 97.307\% & 96.629\% & 95.164\% & 94.355\% & \textbf{100}\% \\
Emot & \textbf{99.872}\% & 87.733\% & 87.071\% & 89.160\% & 69.238\% & 72.912\% & 84.757\% & 98.493\% \\
Gamb & \textbf{96.122}\% & 51.912\% & 59.194\% & 75.553\% & 81.683\% & 68.535\% & 69.023\% & 94.176\% \\
Lang & \textbf{99.159}\% & 88.289\% & 83.235\% & 85.755\% & 89.177\% & 81.517\% & 82.204\% & 97.491\% \\
Moto & \textbf{96.433}\% & 60.449\% & 68.379\% & 77.982\% & 76.858\% & 66.374\% & 77.773\% & 95.754\% \\
Rela & \textbf{98.051}\% & 82.137\% & 75.307\% & 90.609\% & 86.554\% & 87.700\% & 79.628\% & 93.121\% \\
Soci & \textbf{97.544}\% & 55.822\% & 67.123\% & 69.811\% & 73.706\% & 71.455\% & 69.950\% & 96.385\% \\
WMem & \textbf{99.223}\% & 88.274\% & 81.700\% & 78.760\% & 80.202\% & 76.620\% & 82.446\% & 98.299\% \\
\hline
{Average} & \textbf{98.301}\% & 76.827\% & 76.806\% & 83.117\% & 81.756\% & 78.929\% & 80.017\% & 96.710\% \\
\hline
\end{tabular}
}
\begin{tabular}{@{}lll@{}}
CM: Causal Dynamics Model & GNN: Graph Neural Network & DNN: Deep Neural Network~\cite{WX-LX-JZ-NB-ZY-2020} \\
FC: Functional Connectivity & RF: Random Forest & DBRNN: Deep Bidirectional Recurrent Neural Network~\cite{ZH-SH-FA-LO-WU:2022} \\
Raw: Raw data & SVM: Support Vector Machines & BAnD: Brain Attend and Decode~\cite{nguyen2020attend} \\
\end{tabular}
\end{table*}
\subsection{Subject Causal Fingerprinting}
We first verify the use of the proposed Causal-Dynamics Model and Modal Decomposition and Projection (CM+MD\&P) method for subject fingerprinting. It has been well justified in existing literature~\cite{raichle2006brain} that resting-state fMRI exhibits significant variability in activation patterns across different individuals. Our testing data include two resting sessions (Rest1 and Rest2) and two scanning orders (LR and RL). We repeatedly use one out of four possible permutations to construct a labeled database $\mathcal{R}_k$, and use the other three as query data. As shown in Fig. \ref{fig:subject_task-accuracy}-Left, the use of \textit{causal fingerprint} can provide an accuracy of around 80\%, which is notably higher, although causal-based fingerprint quest is a harder classification task, compared to existing methods~\cite{FES-SX-SD-RMD-HJ:15, finn2017can} that use Functional Connectivity for feature extraction and Correlation for fingerprinting (FC+CoR). The comparison of results is obtained by directly applying the methods outlined in Sec. \ref{sec_subjetid} for (CM+MD\&P) and in \cite{FES-SX-SD-RMD-HJ:15} for (FC+CoR) to our dataset, without additional fine-tuning or data processing. Although further processing could enhance the accuracy of both algorithms, it is important to recall that our primary goal in this paper is to verify the feasibility and effectiveness of using causal signatures for fingerprinting. Therefore, we believe that not performing these enhancements allows for a more objective comparison.

A more detailed comparison is provided in Table \ref{tab:my_label}, in which we present the average accuracy (over the cases shown in Fig. \ref{fig:subject_task-accuracy}-Left) by combining different feature extraction (CM/FC) and classification (MD\&P/CoR/FN/GNN) methods, where FN stands for using the Frobenius norm $\|R^Q-R^D\|_F$ to compute the distances between query and database matrices for classification. The proposed CM+MD\&P method outperforms all other combinations. Comparing FC+CoR and FC+MD\&P suggests that the proposed MD\&P can also be used to compare the similarities of the FC matrices and improve fingerprinting accuracy. 
The FN method for classification is not as robust as $MD\&P$.
Additionally, the GNN method is generally not suitable for subject fingerprinting applications due to data availability limitations. Apart from GNN, we also explored other deep-learning models utilizing convolution or attention mechanisms with various layer structures. Similar poor performances are observed.

\subsection{fMRI Task Causal Fingerprinting.}
We verify the use of the proposed Causal Modeling (CM) and GNN method for task fingerprinting as shown in Fig. \ref{fig:subject_task-accuracy}-Right.
Using the causal signatures obtained from the HCP dataset, (50\%) of the data was randomly selected for training and (50\%) for testing.
Notably, the model achieves a perfect accuracy of $100\%$ for the resting state. For Emot, Lang, and WMem, the accuracy is impressively high. On the other hand, Gamb, Moto, Rela, and Soci are lower, and most of the misclassifications occur internally among these tasks.
To demonstrate the advantages of our proposed CM+GNN method, we compared it with several state-of-the-art methods across 8 tasks~\cite{LA-ST-ST-ST-CH:2005,SA-WO-BI-AL-MO:2016,ZH-SH-FA-LO-WU:2022,nguyen2020attend}\footnote{For the experiments that are not open source, we try our best to reproduce the results by following the methods described in the paper.}. The methods include Causal Modeling with Graph Neural Network (CM+GNN), Causal Modeling with Random Forest (CM+RF), Causal Modeling with Support Vector Machine (CM+SVM), Functional Connectivity with Graph Neural Network (FC+GNN), 
and methods using raw data such as Graph Neural Network (Raw+GNN), Deep Neural Network (Raw+DNN), Deep Bidirectional Recurrent Neural Network (Raw+DBRNN), and Brain Attend and Decode (Raw+BAnD).

The result comparison is provided in Table \ref{tab:my_label2}, where we present the average accuracy across various methods. The CM+GNN method proposed in this paper consistently achieves the highest accuracy across multiple tasks, demonstrating its robust capability in task fingerprinting.
The CM+GNN method excels in the most distinctly identifiable task (Rest), achieving perfect accuracy. This high performance is also observed in other methods, indicating that neural patterns in this task are distinctly recognizable. In tasks like emotion (Emot) and language (Lang), CM+GNN shows significant advantages, effectively capturing the unique neural signatures associated with these tasks and surpassing other methods such as CM+RF and Raw+DBRNN.
The performance of CM+GNN slightly degrades in tasks such as gambling (Gamb), motor (Moto), and social (Soci). These tasks likely have more overlapping neural activity patterns, making them harder to distinguish. Nevertheless, note that other methods like Raw+GNN, Raw+DBRNN also have degraded performance in these more challenging tasks. 
We also note that the BAnD method~\cite{nguyen2020attend} uses different data types (volumetric data) and cohort sizes within an end-to-end deep learning scheme. The volumetric data generally has better resolution than the Schaefer-100 parcellation used in our results. Nevertheless, our method achieves comparable results, with even slightly better accuracy.

\subsection{Visualization and Discussions}
\begin{figure}[t]
    \centering
    \includegraphics[width=.48\textwidth]{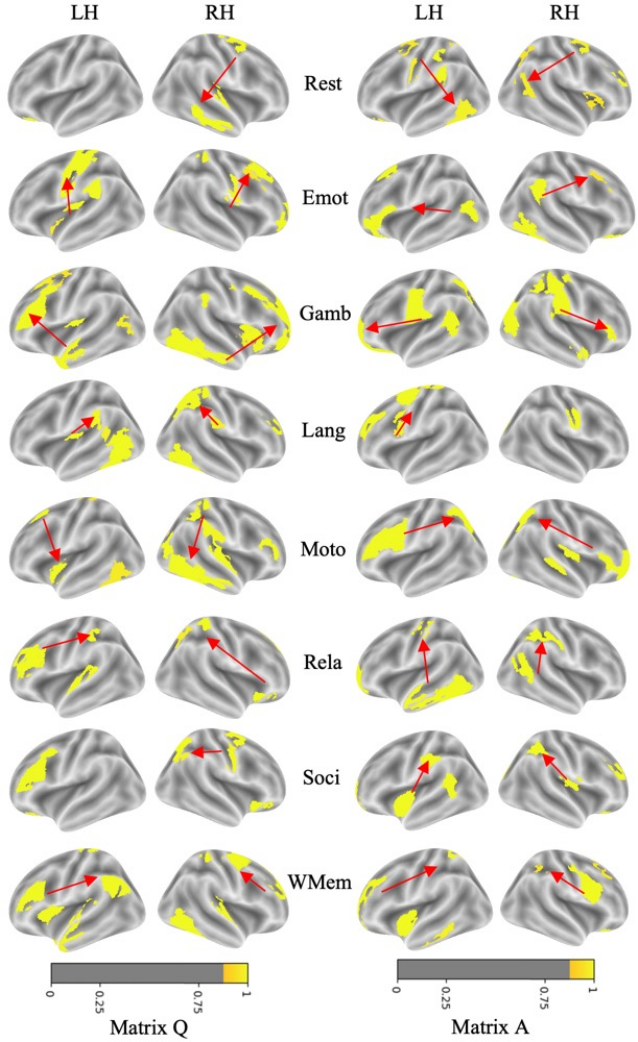}
    \caption{Visualization of Brain Regions with Strongest Connection Weights, Showing Association Strengths Across Multiple Cognitive Tasks (Importance Scores Indicated by Color-Bar). LH refers to the left hemisphere, and RH refers to the right hemisphere of the brain. Arrows indicate the directed relation about which areas cause impacts to other areas. Note that for ease of visualization, we use two colors and set the threshold to highlight only the most active areas of the brain and explain their directional relations.}
    \label{fig:gnn_brain_regions}
\end{figure}

We visualize, in Fig.~\ref{fig:gnn_brain_regions}, the obtained causal signatures for all eight fMRI tasks (Rest, Emot, Gamb, Lang, Moto, Rela, Soci, and WMem)
and analyze their roles in various cognitive and emotional processes. The Matrices $Q$ and $A$ have been normalized for the ease of presenting the results of all cases with a unified metric scale. The entries in matrices $Q$ and $A$ allow us to determine the causal influence direction between different regions based on their numerical values and positions, unlike FC which only shows correlation without specific directional influence. We use arrows in the figures to indicate the directions of these causal influences, which are discussed in detail for each task in the subsequent analysis. By separating matrix $Q$ and matrix $A$, we can observe how brain activity evolves over different timescales. Matrix $Q$ captures fast synchronous interactions among brain regions, reflecting real-time activation patterns during specific tasks. In contrast, Matrix $A$ illustrates slower dynamic evolutions, showing how current states are influenced by previous states. 
The resting state mainly involves the default mode network (DMN), including the medial prefrontal cortex (mPFC) and posterior cingulate cortex (PCC), which remain active in introspective thinking, self-related processing, memory integration and rest-task dichotomy \cite{duong2021morphospace}. The mPFC exerts influence over the PCC, modulating introspective and self-related processing.
The emotion task shows strong connections in the amygdala, prefrontal cortex, and anterior insula, highlighting their key roles in emotion generation and regulation. The amygdala exerts influence over the prefrontal cortex and anterior insula, modulating emotional responses.
The gambling task primarily shows strong connections in the ventral striatum, prefrontal cortex, and insula, which are crucial for reward prediction and decision-making. The ventral striatum influences the prefrontal cortex and insula, affecting decision-making and reward processing.
The language task primarily shows strong activations in Broca’s area (Brodmann areas 44 and 45~\cite{DA-AR-DA-HA:1992}) and Wernicke’s area (Brodmann area 22) in the left hemisphere. Broca’s area is crucial for language production, while Wernicke’s area is essential for language comprehension. Broca’s area influences Wernicke’s area, facilitating coordinated language processing.
The motor task shows strong connections in the primary motor cortex, supplementary motor area (SMA), and basal ganglia, reflecting the importance of these regions in motor planning and execution. The primary motor cortex exerts control over the SMA during motor activities.
The relational task shows strong connections in the dorsolateral prefrontal cortex (DLPFC) and posterior parietal cortex, which play key roles in working memory, abstract thinking, and the formation and retrieval of relational memories. The DLPFC influences the posterior parietal cortex, aiding in complex cognitive processes.
The social task primarily shows strong connections in the medial prefrontal cortex (mPFC) and superior temporal sulcus (STS), indicating the importance of these regions in social cognition and emotional responses. The mPFC influences the STS, modulating social behavior.
The working memory task shows strong connections in the dorsolateral prefrontal cortex (DLPFC), parietal cortex, and anterior cingulate cortex (ACC), showing the core roles of these regions in maintaining and processing information. The DLPFC exerts influence over the parietal cortex, facilitating working memory tasks.

Furthermore, we observe that certain tasks have overlapping strong connections in some brain regions. For example, the gambling, social, and emotion tasks show strong connections in the prefrontal cortex and insula, indicating the collaborative role of these regions in risk decision-making, emotion processing, and social cognition. The language and relational tasks show strong connections in the left temporal regions, particularly Wernicke's area and the hippocampus, showing their crucial roles in language processing and memory integration. The motor and working memory tasks both show strong connections in the dorsolateral prefrontal cortex and parietal cortex, reflecting the importance of these regions in motor planning and working memory.

\section{Conclusion}
In this paper, we verified the feasibility and effectiveness of using fMRI causal signatures for subject and task fingerprints. This was achieved by developing a new causal dynamics representation of fMRI time-series data using a two-timescale linear state-space model. With the spatio-temporal signatures embedded in this representation, we incorporated it with a modal decomposition and projection method to perform subject fingerprinting and a GNN-based model to perform task fingerprinting.
We verified the proposed method using an existing dataset collected from human subjects. The obtained results demonstrated the advantage of our method over alternative methods. 
To the best of our knowledge, we are among the first to investigate and quantify brain large-scale fingerprints in the context of causal dynamics. Through visualizations and discussions, we explored the brain’s dynamic interactions with temporal resolution and captured the directionality of these relationships. This distinguishes our work from conventional fingerprint investigations in brain connectivity domain. In terms of clinical utility, our proposed method holds a potential for the development of diagnostic and prognosis tools in neurodegenerative diseases. 
Along with the proven innovative approach, our investigation has some limitations: i) system inputs are specifically chosen based on different functional areas of the brain, ii) our study only investigated Schaefer $n=100$ parcellation. 
In our future study, we plan to focus on i) investigating different hypothesized system input selections; ii) extending our model to different Schaefer parcellation granularities and atlases; iii) investigating the extension of causal fingerprints in both healthy controls (e.g., other open neuroimaging dataset) and neurodegenerative diseases such as Alzheimer's disease \cite{xu2021optimization,xu2022consistency,garai2024effect}. 

\printbibliography

\appendix

\end{document}